\documentclass[conference]{IEEEtran}
\IEEEoverridecommandlockouts

\usepackage{microtype}
\usepackage{graphicx}
\usepackage[caption=false]{subfig}
\usepackage{booktabs} 
\usepackage{caption}
\usepackage{algorithm}
\usepackage{algpseudocode} 
\usepackage{amsmath}
\usepackage{amssymb}
\usepackage{mathtools}
\usepackage{amsthm}
\usepackage{xspace}
\usepackage[capitalize,noabbrev]{cleveref}
\usepackage{rotating} 
\usepackage{tabularx} 
\usepackage{stfloats}
\theoremstyle{plain}

\theoremstyle{definition}

\theoremstyle{remark}

\usepackage[textsize=tiny]{todonotes}

\definecolor{darkred}{RGB}{150,0,0}
\definecolor{darkgreen}{RGB}{0,150,0}
\definecolor{darkblue}{RGB}{0,0,150}
\usepackage{graphicx,bm,epsfig,epsf,color,mathbbol,fmtcount,semtrans,multirow,comment,boldline,pgfplots}
\usepackage{tikz}
\usetikzlibrary{pgfplots.groupplots}
\usepackage[utf8]{inputenc} 
\usepackage[T1]{fontenc}    
\usepackage{url}            
\usepackage{amsfonts}       
\usepackage{nicefrac}       

\newcommand{\cln}[1]{\textcolor{red}{}}

\newcommand{\confed}{\textsc{FLASH}\xspace}
\newcommand{\confedmab}{\textsc{TSscores}}

\newcommand{\beq}{\begin{equation}}
\newcommand{\ba}{\begin{align}}
\newcommand{\ea}{\end{align}}

\newcommand{\eeq}{\end{equation}}

\newcommand{\V}{{\mtx{V}}}

\newcommand{\bt}{{\boldsymbol{\theta}}}

\newcommand{\Sc}{\boldsymbol{\mathcal{S}}}

\newcommand{\w}{\vct{w}}

\newcommand{\x}{\vct{x}}

\newcommand{\rb}{\vct{r}}
\newcommand{\rba}{\vct{r}^{\text{all}}}

\newenvironment{myitemize}{\begin{list}{$\bullet$}
		{\setlength{\topsep}{1mm}
			\setlength{\itemsep}{0.25mm}
			\setlength{\parsep}{0.25mm}
			\setlength{\itemindent}{0mm}
			\setlength{\partopsep}{0mm}
			\setlength{\labelwidth}{15mm}
			\setlength{\leftmargin}{4mm}}}{\end{list}}

\definecolor{emmanuel}{RGB}{255,127,0}

\newcommand{\R}{\mathbb{R}}

\newcommand{\vct}[1]{\bm{#1}}
\newcommand{\mtx}[1]{\bm{#1}}

\newcommand{\X}{{\mtx{X}}}
\newcommand{\Xa}{\mtx{X}^{\text{all}}}
\newcommand{\Xat}{\mtx{X}^{\text{all}^{\top}}}

\newlength{\commentWidth}
\usepackage{cite}
\usepackage{amsmath,amssymb,amsfonts}
\usepackage{graphicx}
\usepackage{textcomp}
\usepackage{xcolor}
\def\BibTeX{{\rm B\kern-.05em{\sc i\kern-.025em b}\kern-.08em
    T\kern-.1667em\lower.7ex\hbox{E}\kern-.125emX}}
     \pgfplotsset{compat=1.18}

\begin{document}

\title{FLASH: Federated Learning Across Simultaneous Heterogeneities}

\author{
  \IEEEauthorblockN{
    Xiangyu Chang\textsuperscript{1},
    Sk Miraj Ahmed\textsuperscript{1,2},
    Srikanth V. Krishnamurthy\textsuperscript{1},
    Basak Guler\textsuperscript{1},
    Ananthram Swami\textsuperscript{3},\\
    Samet Oymak\textsuperscript{4},
    Amit K. Roy-Chowdhury\textsuperscript{1}
  }
  \IEEEauthorblockA{\textsuperscript{1}University of California, Riverside, \textsuperscript{2}Brookhaven National Laboratory}
  \IEEEauthorblockA{\textsuperscript{3}DEVCOM Army Research Laboratory, \textsuperscript{4}University of Michigan}
  \IEEEauthorblockA{Email: \texttt{\textsuperscript{1}\{cxian008, krish@cs, basakg, amitrc@ece\}@ucr.edu, \textsuperscript{2}sahme047@ucr.edu,}\\
    \texttt{\textsuperscript{3}ananthram.swami.civ@army.mil,
    \textsuperscript{4}oymak@umich.edu}
  }
}

\maketitle

\begin{abstract}

The key premise of federated learning (FL) is to train ML models across a diverse set of data-owners (clients), without exchanging local data. An overarching challenge to this date is client heterogeneity, which may arise not only from variations in data distribution, but also in data quality, as well as compute/communication latency. An integrated view of these diverse and concurrent sources of heterogeneity is critical; for instance, low-latency clients may have poor data quality, and vice versa. In this work, we propose \confed~(\underline{F}ederated \underline{L}earning \underline{A}cross \underline{S}imultaneous \underline{H}eterogeneities), a lightweight and flexible \emph{client selection} algorithm that outperforms state-of-the-art FL frameworks under extensive sources of heterogeneity, by trading-off the statistical information associated with the client's data quality, data distribution, and latency. \confed~is the first method, to our knowledge, for handling all these heterogeneities in a unified manner. To do so, \confed~models the learning dynamics through contextual multi-armed bandits (CMAB) and dynamically selects the most promising clients. Through extensive experiments, we demonstrate that \confed~achieves substantial and consistent improvements over state-of-the-art baselines---as much as 10\% in absolute accuracy---thanks to its unified approach. Importantly, \confed~also outperforms federated aggregation methods that are designed to handle highly heterogeneous settings and even enjoys a performance boost when integrated with them.\\    
\end{abstract}

\section{Introduction}

Federated Learning (FL) is a distributed learning paradigm where multiple clients collaborate to train a model without exchanging raw data. Training is coordinated by a central server, who selects clients \cite{mcmahan2017communication,lai2021oort,li2022pyramidfl,yemini2021restless,cao2022birds,FedCor,ren2020scheduling} in each round to update the aggregated global model\cite{mcmahan2017communication,FedProx,FedBiO,FedDF,wang2020tackling,karimireddy2020scaffold,RHFL}. While FL offers advantages in privacy, reduced communication costs, and scalability, it faces unique challenges due to its distributed nature, particularly in handling client heterogeneity, ensuring fairness and robustness, and balancing model accuracy with privacy.\\
Client heterogeneity is a central challenge in FL, manifesting through non-homogeneous label distribution~\cite{FedProx}, unreliable label assignment~\cite{FedCor}, and latency~\cite{li2022pyramidfl}. This heterogeneity degrades model accuracy \cite{mcmahan2017communication} and increases training resources required. While existing research suggests that informed client selection can address these issues\cite{mcmahan2017communication,lai2021oort,li2022pyramidfl,yemini2021restless,cao2022birds,FedCor,ren2020scheduling}, current methods typically handle only one or two types of heterogeneities \cite{li2022pyramidfl}. This limitation, combined with the need for client selection algorithms that integrate seamlessly with federated aggregation strategies, motivates our research question:

\begin{figure}[t]
    \centering
    \hspace*{-15pt}\includegraphics[width=8cm]{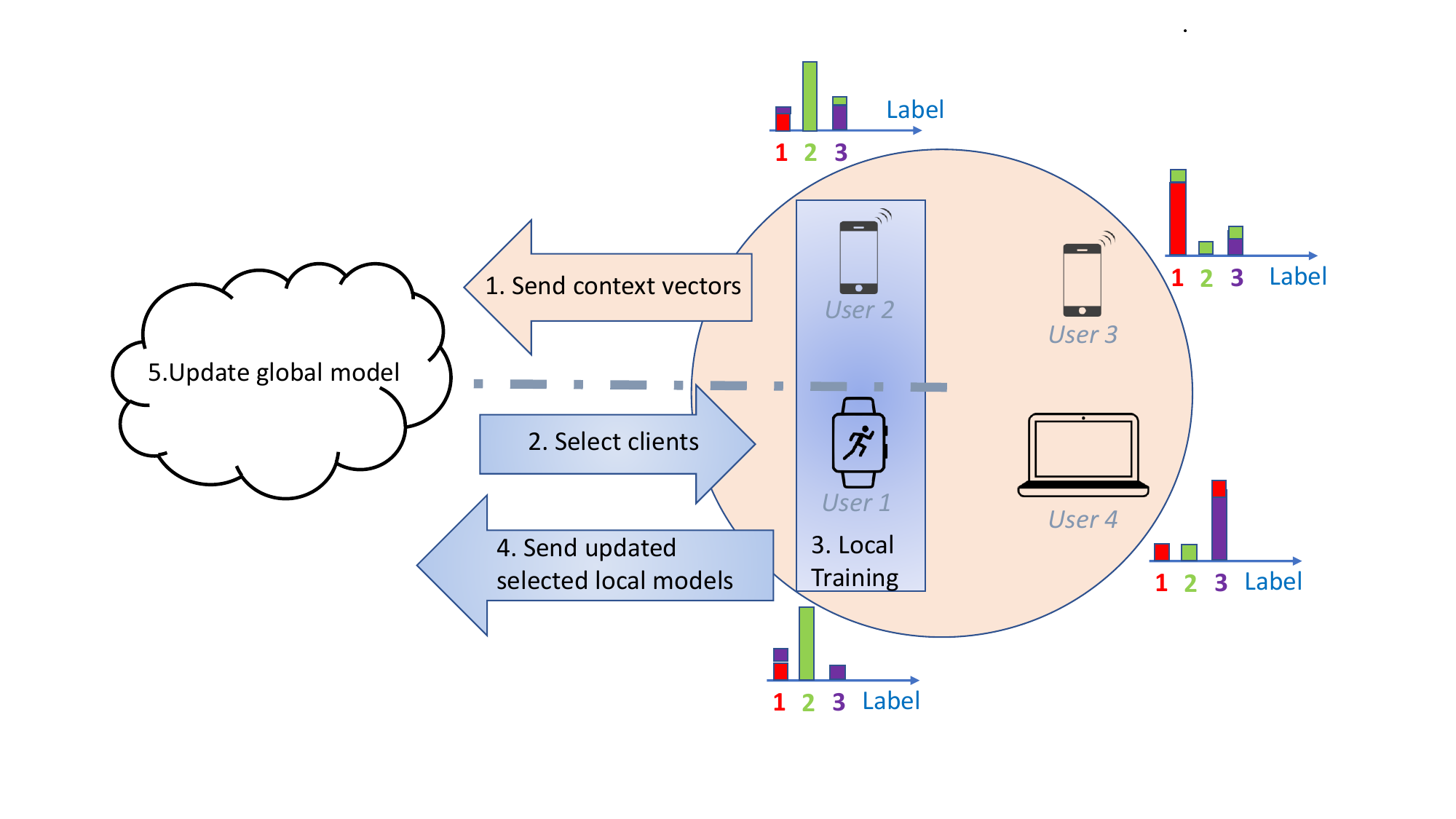}\vspace{-15pt}
    \caption{Problem setup for \confed: Building upon the standard federated learning setup of a global model learned from updates from local clients, we consider the setting where the labels of the data at the clients are imprecise (mismatched colors for the labels at each client indicate noise in those labels), the distribution of the data classes across the clients is non-uniform (height of the bars for each label class at each client), and the latencies of clients are variable (e.g. diverse devices, varying communication distance, etc.). We term these variations as heterogeneities in the data. \confed~is built upon a contextual multi-armed bandit approach, which selects the optimal set of users to update the global model, where the context vectors of clients capture their various heterogeneities. The main steps (1-4) of \confed~are illustrated in the figure.} 
    \label{fig:overview}
\end{figure}
\emph{\textbf{Q:} How can we select clients systematically under diverse and concurrent sources of heterogeneity to facilitate faster training and better accuracy? Can we combine the benefits of our client selection method with existing aggregation methods?}

\noindent 
\textbf{Main contribution.} Our algorithm \confed~addresses this challenge by explicitly modeling client heterogeneity as a context vector that summarizes client characteristics, including distributional heterogeneity, label noise, and straggler latency. \confed~employs Contextual Multi-Armed Bandits (CMAB) to select clients that maximize improvement in the global optimization objective. It offers three key advantages:\\
$\bullet$ \textbf{Simultaneous and diverse heterogeneities:} \confed~uniquely addresses multiple concurrent sources of heterogeneity through contextual variables, facilitating optimal trade-offs between diverse heterogeneities in complex scenarios.\\
$\bullet$ \textbf{Contextual and interpretable framework:} The novel CMAB framework employs contextual features to predict client contributions to global accuracy, with ablation studies confirming its ability to emphasize relevant features as heterogeneity changes.\\
$\bullet$ \textbf{Significant Performance Improvement:} \confed~achieves up to \textbf{10\% improvement} in accuracy over state-of-the-art baselines and demonstrates superior performance when combined with various federated learning aggregation methods.

\section{Methodology: \confed~Algorithm}
\label{sec:approach}

Our goal is to efficiently select clients to train a global model subject to multiple sources of client heterogeneity. The key idea is that the central server uses each client's contextual information to select those that most improve the global model's accuracy. Since the server lacks a priori knowledge of client latencies, data diversity, or label noise, it must dynamically select the most contributive clients, as shown in Fig.~\ref{fig:overview}. Multi-Armed Bandits (MAB)\cite{katehakis1995sequential} provide an effective framework for such decision-making problems.\\
The client selection policy must be determined concurrently with training, which makes approaches with higher sample complexity infeasible. Experimentally, we find that more complex approaches (e.g., using neural net-based bandits) can slow down optimization and harm accuracy (see Fig \ref{fig:ablation_and_bandit}). This motivates our choice of a sample-efficient contextual MAB (CMAB)\cite{qin2014contextual} framework that rapidly adapts to federated optimization dynamics by incorporating client heterogeneity statistics and prior rewards in its context vector (see Algo.~\ref{alg:confeddi1}).

\subsection{Contextual MAB with Thompson-Sampling}
\label{sec:MAB}
Let $[m]:=\{1,2,\dots,m\}$ denote the set of clients and ${\Sc_t}$ be the set of all feasible subsets in round $t$. At round $t$, a learner observes $m$ $d$-dimensional context vectors $\left\{ {{\x_t}( 1 ),\dots,{\x_t}( m )} \right\} \subseteq {\mathbb{R}^d}$ corresponding to the $m$ arms (Algorithm \ref{alg:confeddi1}, lines 3-4). The learner chooses a super arm ${S_t} \in {\Sc_t}$ containing $M_t$ clients and observes rewards ${\rb_t=\left\{ {r_t(i)} \right\}_{i \in {S_t}}}$, receiving total reward ${R_t}( {{S_t}} )=\sum_{i\in S_t} r_t(i)$. For linear bandits, the expected reward follows: $\mathbb{E}\left[ {\left. {{r_t}(i)}~\right|~{\x_t}(i)} \right] = \bt _*^\top{\x_t}( i ).$\\
\textbf{Thompson Sampling Procedure (Algorithm \ref{alg:confeddi2}).} Using previous selections' history $\{(S_\tau)_{\tau=1}^{t-1},\Xa_{t-1},\rba_{t-1}\}$, we estimate $\bt_*$ via ridge regression (Algorithm \ref{alg:confeddi2}, lines 2-3):
$\hat\bt_{t}=\V_t^{-1}\Xat_{t-1}{{\rba_{t-1}}}~~\text{where}~~
\V_t={\Xat_{t-1}{\Xa_{t-1}}{\text{ + }}\lambda {\mathbf{I}}}.$
Thompson Sampling models the distribution of $\bt$ at time $t$ as $\mathcal{N}(\hat\bt_t,\gamma_t^2\V_t^{-1})$ with ${\gamma _{t}} = {\lambda ^{1/2}} + \sqrt {d\ln \left( {\frac{{1 + tm}}{\delta }} \right)}$, drawing a sample $\hat\bt_{\text{new}}$ to score clients by expected rewards $\hat{r}_t(i)=\x_t(i)^\top \hat\bt_{\text{new}}$ (Algorithm \ref{alg:confeddi2}, lines 4-6). For computational efficiency, $\V_t$ is updated incrementally as $\V_{t+1} \leftarrow \V_t + \sum_{i\in S_t}\x_i\x_i^\top$ (Algorithm \ref{alg:confeddi1}, lines 13-14).

\noindent\textbf{Properties of~\confed.} The selected clients influence the global model, causing both rewards and context vectors to change in subsequent rounds (Algorithm \ref{alg:confeddi1}, lines 8-12). As \confed~samples a client more frequently, the client's context vector changes, making it more likely that a different client will be selected in future rounds, preventing repeatedly choosing the same clients.

\subsection{Noise Robust Training}
\label{sec:noiserobusttraining}

\begin{algorithm}[t!]
\caption{\pmb{\confed:} Heterogeneity-aware Client Selection}
\label{alg:confeddi1}
\begin{algorithmic}[1]
\State \textbf{Input:} Initial model $\w^0$, Number of FL rounds $n$, Number of clients $m$, Split local dataset $\{\mathcal{D}_i\}_{i=1}^m$ into local training and validation sets $[\mathcal{T}_i;\mathcal{V}_i]_{i=1}^m$, Number of clients to select $(M_t)_{t=0}^n$, Exploration strength $\gamma_t\geq 0$, Regularization strength $\lambda$, Confidence $\delta$
\State \textbf{Output:} Final model $\w^n$
\State Initialize MAB parameter estimation $\hat\bt_0\gets 0$
\State $S_0\gets[m]$
\State $\vct{b}_0\gets0$
\State $\V_0\gets\lambda \vct{I}_d$
\For {rounds $t=0,1,...,n-1$}
    \State $\gamma_t={\lambda ^{1/2}} + \sqrt {d\ln \left( {\frac{{1 + tm}}{\delta }} \right)} $
    \State \textbf{Server:} Send $\w^t$ to all clients $i \in [m]$
    \For {client $i\in S_t$}
        \State Download global model $\w^t$
        \State $\w_i^{t+1} \gets\texttt{LocalTraining}(\w^t,\mathcal{T}_i)$
    \EndFor
    \For {all clients $i\in [m]$}
        \State Measure the duration $\tau$
        \State  $\x_{t}(i)\gets \texttt{GetContext}(\mathcal{V}_i,\mathcal{T}_i,\tau)$
    \EndFor
    \State \emph{FedAvg}: $\w^{t+1} \gets \sum_{i \in {S_t}} ({{N_i}}/{N}) \w_i^{t + 1}$
    \Statex {// $\X_{t}\gets$ Concatenate $ (\x_{t}(i))_{i\in S_t}$}
    \State $\rb_{t}\gets \texttt{GlobalModelEvaluation}(\mathcal{V},\w^{t})$
    \Statex {// Concatenate $\Xa_{t}\gets [\X_{t},\Xa_t]$}
    \Statex {// Concatenate $\rba_{t}\gets [\rb_{t},\rba_t]$}
    \State $\texttt{scores}_t, \V_{t+1},\vct{b}_{t+1}\gets$ \Statex \confedmab$(\{\x_i\}_{i\in [m]},\{r_i\}_{i\in S_t},\V_t,\vct{b}_t,\gamma_t, S_t)$
    \State $S_{t+1} \gets \text{top\_}M_{t+1}\text{\_indices}(\texttt{scores}_t)$

\EndFor\\
\Return Final model $\w^n$
\State // Final model is evaluated on a global test dataset $\mathcal{G}$
\end{algorithmic}
\end{algorithm}

\begin{algorithm}[t!]
\caption{\pmb{\confedmab:} Thompson Sampling-based client scores}
\label{alg:confeddi2}
\begin{algorithmic}[1]
\State \textbf{Input:} Data $\{\x_i\}_{i\in [m]},\{r_i\}_{i\in S_t}$, current parameter $\V,\vct{b}$, exploration strength $\gamma_t\geq 0$, selected clients $S_t$

\State \textbf{Output:} Client selection $\texttt{scores}\in\R^m$
\State // Equivalent to $\V \gets {\Xa}^\top{\Xa}+\lambda \vct{I}_d$
\State $\V \gets \V + \sum_{i\in S_t}\x_i\x_i^\top$
\State $\vct{b} \gets \vct{b}+\sum_{i\in S_t}r_i\x_i$
\State $\hat\bt \gets \V^{-1} \vct{b}$
\State $\hat \bt_{new} \text{ is sampled from } \mathcal{N}(\hat\bt,{\gamma_t^2}{} {\V^{-1}})$

\For {all clients $i\in [m]$} 
\State $\begin{gathered}\texttt{scores}(i) \gets \hat \bt_{new} ^\top{\x}\left( i \right)\end{gathered}$
    \EndFor\\
\Return $\texttt{scores}, \V,\vct{b}$
\end{algorithmic}
\end{algorithm}
To reduce the impact of noisy labels, \confed~employs pseudo-labeling techniques and measures model performance on both actual and pseudo-labels. For a $K$-class classification problem with dataset $\mathcal{D} = \{\vct{a}^i, \vct{y}^i\}_{i=1}^n$, we aim to learn a model $p(\cdot; \vct{w}^t):\mathcal{X}\rightarrow\mathcal{Y}$. The local dataset $\mathcal{D}_{i}$ is split into training set $\mathcal{T}_i$ and validation set $\mathcal{V}_i$. We generate soft pseudo-labels using model output $\vct{z}:= p(\vct{a};\vct{w}^t)$ for samples from $\mathcal{T}_i$ (Algorithm \ref{alg:confeddi1}, line 7).

The regular cross-entropy loss $\mathcal{L}_{CE}$ and reverse cross-entropy loss $\mathcal{L}_{RCE}$ are:
\begin{align} 
\mathcal{L}_{CE}(\mathcal{D})&=-\frac{1}{|\mathcal{D}|}\sum_{(\vct{a}, \vct{y})\in\mathcal{D}}\sum_{k=1}^K  [\vct{y}]_k\log[p(\vct{a};\vct{w}^t)]_k,\\
\mathcal{L}_{RCE}(\mathcal{D})&=-\frac{1}{|\mathcal{D}|}\sum_{(\vct{a}, \vct{y})\in\mathcal{D}}\sum_{k=1}^K [p(\vct{a};\vct{w}^t)]_k \log [\vct{y}]_k 
\end{align}

The noise-robust loss combines these losses:
\begin{equation}
    \begin{aligned}
    \mathcal{L}_{robust}(\mathcal{D}_i)&=\mathcal{L}_{CE}(\mathcal{T}_i)+\alpha\mathcal{L}_{CE}(\mathcal{P}_i)+\beta\mathcal{L}_{RCE}(\mathcal{T}_i)
    \end{aligned}
\end{equation}
where $\alpha$ controls overfitting and $\beta$ allows flexible exploration of RCE robustness. For computational stability with one-hot labels, we define $\log 0=A$ where $A<0$ is a constant. 

\subsection{Reward and Context Vector for \confed}
\label{sec:context}
The reward is defined as the average pseudo-label CE loss change rate: $r_{t}=\frac{|\mathcal{L}_{robust}^{t}-\mathcal{L}_{robust}^{t-1}|}{\tau_{t-1}}$ (Algorithm \ref{alg:confeddi1}, line 11). The context vector for client $i$ in round $t$ is $\x_t(i)=[\mathcal{L}_{robust}^t/\mathcal{L}_{robust}^1, \mathcal{L}_{CE}^t(\mathcal{V}_i)/\mathcal{L}_{CE}^1(\mathcal{V}_i), \tau_{t-1}, r_{t-1}]$ (Algorithm \ref{alg:confeddi1}, line 3), incorporating:

\begin{myitemize}
\item \textbf{Local training loss ratio} $\mathcal{L}_{robust}^t/\mathcal{L}_{robust}^1$: Normalized training loss reflecting relative change
\item \textbf{Local validation loss ratio} $\mathcal{L}_{CE}^t(\mathcal{V}_i)/\mathcal{L}_{CE}^1(\mathcal{V}_i)$: Measure of overfitting and data heterogeneity
\item \textbf{Duration} $\tau_{t-1}$: Computation time, simulated using shifted-exponential distribution
\item \textbf{Previous reward} $r_{t-1}$: Included for more accurate prediction of reward changes.
\end{myitemize}

\section{Theoretical Analysis of Linear Contextual Bandits for Client Selection}

\subsection{Modeling Assumptions for Regret Analysis}
For the purpose of the regret analysis presented herein, we adopt the following standard modeling assumptions:
\begin{enumerate}
    \item \textbf{Linear Expected Reward}: The expected reward $\mathbb{E}[r_t(i) | \x_t(i)]$ for any client $i \in [m]$ (where $m$ is the total number of clients) given its $d$-dimensional context vector $\x_t(i) \in \mathbb{R}^d$ at round $t \in [0, n-1]$ (where $n$ is the total number of FL rounds) is assumed to be linear. This is modeled as $\mathbb{E}[r_t(i) | \x_t(i)] = \theta^{*\top} \x_t(i)$, where $\theta_i^*$ is an unknown true $d$-dimensional parameter vector specific to client $i$. This per-client parameterization allows for a fine-grained analysis within this theoretical framework.

    \item \textbf{Conditionally $R$-sub-Gaussian Noise}: The observed reward $r_t(i)$ from client $i$ at round $t$ is $r_t(i) = \theta^{*\top} \x_t(i) + \eta_t(i)$. The noise term $\eta_t(i)$ is assumed to be conditionally $R$-sub-Gaussian,, for some constant $R > 0$.

    \item \textbf{Bounded Norms}: The $L_2$ norms of the context vectors and the true parameter vectors are assumed to be bounded. That is, for all clients $i$ and rounds $t$, $||\x_t(i)||_2 \le L$ and $||\theta_i^*||_2 \le S$, for some positive constants $L$ and $S$.
\end{enumerate}

\subsection{Cumulative Regret Definition}
The performance of the client selection algorithm is evaluated using cumulative regret, denoted $\mathcal{R}_n$. Let $S_t^*$ be the optimal set of $M_t$ clients that would be chosen at round $t$ by an oracle with full knowledge of all true parameter vectors $\theta_i^*$. Let $S_t$ be the set of $M_t$ clients selected by the learning algorithm at round $t$. The cumulative regret over $n$ rounds is:
$$ \mathcal{R}_n = \sum_{t=0}^{n-1} \left[ \sum_{i \in S_t^*} \mathbb{E}[r_t(i)|\x_t(i)] - \sum_{i \in S_t} \mathbb{E}[r_t(i)|\x_t(i)] \right] $$

\subsection{Regret Bound for a LinUCB-type Selector}
Consider a LinUCB-type client selection algorithm operating under the assumptions outlined above. If the exploration parameter $\alpha_t$ used for computing the Upper Confidence Bound (UCB) for client scores at round $t \in [0, n-1]$ is defined as:
$\gamma_t = R\sqrt{d\log\left(\frac{1+tL^2/\lambda}{\delta}\right)} + \lambda^{1/2}S$, 
where $d$ is the dimension of the context vectors $\x_t(i)$, $\lambda > 0$ is the regularization strength (as used in $V_0 = \lambda I_d$ within the \confed MAB initialization), and $\delta \in (0,1)$ is a confidence parameter (an input to \confed, used in Alg.~\ref{alg:confeddi1} for $\gamma_t$).

Then, for a strategy that selects $M_t$ clients per round (where $M_t$ is the number of clients to select, an input to \confed) over $n$ total communication rounds, the cumulative regret $\mathcal{R}_n$ from Alg.~\ref{alg:confeddi1} can be proven (similarly as in \cite{ qin2014contextual}) to be upper bounded by (with probability $1-\delta$):
\begin{align*}
    \mathcal{R}_n\le &\gamma_n\|\x_n(i)\|_{\V_n^{-1}} \le M_n \log\left(1+\frac{nL^2}{5\lambda}\right)\\&\cdot\left(R\sqrt{d\log\left(\frac{1+nL^2/\lambda}{\delta}\right)} + \lambda^{1/2}S\right)
\end{align*}
Where $\V_n^{-1}$ comes from Alg.~\ref{alg:confeddi2}. This bound indicates that the regret grows sub-linearly with the number of rounds $n$, demonstrating the learning capability of such an approach.
\section{Experimental Evaluation}
\begin{table*}[htbp]
  \centering
  \title{30\% non-IIDness}
 \resizebox{0.7\textwidth}{!}{ %
  \begin{tabular}{c ccccccc c}

    \toprule
    & FedAvg \cite{mcmahan2017communication} & FedProx \cite{FedProx} & FedBiO \cite{FedBiO} & FedDF \cite{FedDF} & FedNova \cite{wang2020tackling} & SCAFFOLD \cite{karimireddy2020scaffold} & RHFL \cite{RHFL} & Average \\
    \cmidrule(lr){2-8} \cmidrule(lr){9-9}
    Random \cite{mcmahan2017communication}  & 57.1 $\pm$ 2.4 & 64.3 $\pm$ 2.5 & 65.2 $\pm$ 2.7 & 65.5 $\pm$ 3.5 & 67.8 $\pm$ 3.2 & 69.3 $\pm$ 2.6 & 64.7 $\pm$ 2.7 & 64.8 \\
    Oort \cite{lai2021oort} & 65.3 $\pm$ 1.6 & 66.5 $\pm$ 1.4 & 67.2 $\pm$ 1.1 & 70.1 $\pm$ 1.2 & 67.2 $\pm$ 1.1 & 69.0 $\pm$ 1.4 & 65.2 $\pm$ 1.4 & 67.2\\
    PyramidFL \cite{li2022pyramidfl} & 67.1 $\pm$ 1.1 & 68.7 $\pm$ 1.0 & 69.1 $\pm$ 1.3 & 68.0 $\pm$ 1.5 & 69.3 $\pm$ 1.3 & 70.1 $\pm$ 1.6 & \underline{66.5 $\pm$ 1.2} & 68.4\\
    Restless bandit \cite{yemini2021restless}  & 66.6 $\pm$ 2.9 & 63.2 $\pm$ 2.2 & 66.1 $\pm$ 2.5 & 67.6 $\pm$ 2.4 & 62.4 $\pm$ 2.3 & 63.2 $\pm$ 1.6 & 64.3 $\pm$ 2.0 & 64.7\\
    Neural bandit \cite{cao2022birds}  & 69.3 $\pm$ 2.6 & 71.2 $\pm$ 1.8 & 69.3 $\pm$ 2.3 & \underline{70.4 $\pm$ 2.4} & 69.0 $\pm$ 2.1 & 71.8 $\pm$ 2.2 & 65.9 $\pm$ 2.5 & 69.6\\
    FedCor \cite{FedCor} & \underline{70.1 $\pm$ 1.4} & \fbox{73.2 $\pm$ 1.5} & \fbox{73.8 $\pm$ 1.2} & 69.5 $\pm$ 1.6 & \underline{71.8 $\pm$ 1.6} & \underline{72.8 $\pm$ 1.4} & 64.8 $\pm$ 1.6& \underline{70.8}\\
    FEEL \cite{ren2020scheduling} & 65.5 $\pm$ 1.1& 69.4 $\pm$ 1.5 & 67.3 $\pm$ 1.6 & 69.7 $\pm$ 1.3 & 70.4 $\pm$ 1.4 & 64.6 $\pm$ 1.2 & 65.3 $\pm$ 1.5& 67.4\\
    \confed (Ours)  & \fbox{70.3 $\pm$ 1.1} & \underline{71.6 $\pm$ 1.3} & \underline{72.7 $\pm$ 1.2} & \fbox{72.2 $\pm$ 1.1} & \fbox{73.5 $\pm$ 1.4} & \fbox{73.7 $\pm$ 1.3} & \fbox{69.2 $\pm$ 1.3} & \fbox{71.8}\\
    \bottomrule
  \end{tabular}}
  \centering
\title{15\% label noise}
  \resizebox{0.7\textwidth}{!}{\begin{tabular}{c ccccccc c}
    \toprule
    & FedAvg \cite{mcmahan2017communication} & FedProx \cite{FedProx} & FedBiO \cite{FedBiO} & FedDF \cite{FedDF} & FedNova \cite{wang2020tackling} & SCAFFOLD \cite{karimireddy2020scaffold} & RHFL \cite{RHFL}  & Average \\
    \cmidrule(lr){2-8} \cmidrule(lr){9-9}
    Random \cite{mcmahan2017communication} & 57.6 $\pm$ 2.7 & 61.1 $\pm$ 3.1 & 61.3 $\pm$ 3.4 & 60.3 $\pm$ 2.5 & 59.6 $\pm$ 3.1 & 57.2 $\pm$ 2.2 & 68.3 $\pm$ 2.4 & 60.8\\
    Oort \cite{lai2021oort} & 61.4 $\pm$ 1.8 & 66.8 $\pm$ 1.4 & 66.4 $\pm$ 1.2 & \underline{67.6 $\pm$ 1.3} & 65.8 $\pm$ 1.2 & 66.3 $\pm$ 1.5 & 69.4 $\pm$ 1.4 & 66.3\\
    PyramidFL \cite{li2022pyramidfl} & 63.9 $\pm$ 1.2 & 66.1 $\pm$ 1.1 & 65.7 $\pm$ 1.2 & 64.5 $\pm$ 1.4 & 65.7 $\pm$ 1.4 & 66.6 $\pm$ 1.6 & 70.0 $\pm$ 1.3 & 66.0\\
    Restless bandit \cite{yemini2021restless} & 60.5 $\pm$ 3.0 & 60.7 $\pm$ 2.6 & 64.1 $\pm$ 2.4 & 58.9 $\pm$ 2.3 & 54.3 $\pm$ 1.9 & 60.2 $\pm$ 1.6 & 69.0 $\pm$ 2.1 & 61.1\\
    Neural bandit \cite{cao2022birds} & 63.6 $\pm$ 2.4 & 68.7 $\pm$ 2.2 & \underline{66.8 $\pm$ 2.4} & 63.4 $\pm$ 2.1 & 67.6 $\pm$ 2.0 & 68.2 $\pm$ 1.7 & 71.4 $\pm$ 2.5 & 67.1\\
    FedCor \cite{FedCor} & \underline{66.6 $\pm$ 1.5} & \fbox{71.4 $\pm$ 1.3} & 66.4 $\pm$ 1.2 & 63.6 $\pm$ 1.8 & \underline{67.9 $\pm$ 1.4} & \underline{70.3 $\pm$ 1.7} & 69.1 $\pm$ 1.8 & \underline{67.9}\\
    FEEL \cite{ren2020scheduling} & 62.5 $\pm$ 1.4 & 64.9 $\pm$ 1.7 & 59.2 $\pm$ 1.5 & 67.0 $\pm$ 1.6 & 65.5 $\pm$ 2.1 & 62.3 $\pm$ 2.3 & \underline{71.5 $\pm$ 1.5} & 64.7\\
    \confed (Ours) & \fbox{68.2 $\pm$ 1.2} & \underline{69.0 $\pm$ 1.4} & \fbox{71.5 $\pm$ 1.4} & \fbox{72.2 $\pm$ 1.2} & \fbox{69.8 $\pm$ 1.3} & \fbox{71.9 $\pm$ 1.4} & \fbox{74.2 $\pm$ 1.5} & \fbox{70.9}\\
    \bottomrule
  \end{tabular}}
    \resizebox{0.7\textwidth}{!}{\begin{tabular}{c ccccccc c}
    \toprule
    & FedAvg\cite{mcmahan2017communication} & FedProx\cite{FedProx} & FedBiO\cite{FedBiO} & FedDF\cite{FedDF} & FedNova\cite{wang2020tackling} & SCAFFOLD\cite{karimireddy2020scaffold} & RHFL\cite{RHFL}  & Avg \\
    \cmidrule(lr){2-8} \cmidrule(lr){9-9}
Random\cite{mcmahan2017communication} & 47.6 $\pm$ 3.9 & 49.8 $\pm$ 2.7 & 50.7 $\pm$ 3.1 & 49.2 $\pm$ 2.6 & 48.1 $\pm$ 3.0 & 46.1 $\pm$ 2.8 & 57.0 $\pm$ 2.4 & 49.8\\
Oort\cite{lai2021oort} & 51.2 $\pm$ 2.5 & 54.7 $\pm$ 2.3 & 56.3 $\pm$ 2.1 & \underline{56.9 $\pm$ 2.4} & 55.5 $\pm$ 2.2 & 56.0 $\pm$ 2.2 & 57.6 $\pm$ 2.5 & 55.4\\
PyramidFL\cite{li2022pyramidfl} & 52.4 $\pm$ 2.6 & 54.4 $\pm$ 2.3 & 53.9 $\pm$ 2.2 & 52.5 $\pm$ 2.0 & 53.9 $\pm$ 2.3 & 54.9 $\pm$ 2.1 & 58.1 $\pm$ 2.1 & 54.3\\
Restless bandit\cite{yemini2021restless} & 46.2 $\pm$ 3.5 & 49.4 $\pm$ 2.9 & 52.6 $\pm$ 3.2 & 47.8 $\pm$ 3.1 & 49.3 $\pm$ 3.1 & 48.8 $\pm$ 2.7 & 55.3 $\pm$ 2.9 & 49.9\\
Neural bandit\cite{cao2022birds} & \underline{52.6 $\pm$ 2.8} & 53.0 $\pm$ 2.6 & 54.3 $\pm$ 2.5 & 52.0 $\pm$ 2.2 & 56.2 $\pm$ 2.4 & 56.8 $\pm$ 2.4 & 56.9 $\pm$ 2.3 & 54.5\\
FedCor\cite{FedCor} & 51.7 $\pm$ 1.9 & \underline{54.9 $\pm$ 2.1} & \underline{58.0 $\pm$ 2.3} & 52.7 $\pm$ 2.4 & \underline{56.4 $\pm$ 2.1} & \underline{58.8 $\pm$ 2.3} & 57.7 $\pm$ 1.9 & \underline{55.9}\\
FEEL\cite{ren2020scheduling} & 51.2 $\pm$ 2.5 & 53.4 $\pm$ 2.1 & 57.7 $\pm$ 2.2 & 55.5 $\pm$ 2.0 & 54.0 $\pm$ 2.1 & 50.8 $\pm$ 2.0 & \underline{59.7 $\pm$ 2,3} & 54.6\\
\confed (Ours) & \fbox{56.4 $\pm$ 2.4} & \fbox{57.6 $\pm$ 2.1} & \fbox{60.0 $\pm$ 1.8} & \fbox{60.8 $\pm$ 2.2} & \fbox{58.3 $\pm$ 2.1} & \fbox{60.4 $\pm$ 1.7} & \fbox{62.7 $\pm$ 1.6} & \fbox{59.4}\\
    \bottomrule
\end{tabular}}
  \caption{Combining selection-based algorithms with aggregation-based algorithms, we compare the accuracy (in \%) of \confed~on CIFAR-10 dataset. Upper: 30\% non-IIDness, Middle: 15\% label noise, Lower: 30\% non-IIDness and 15\% label noise. Rows are client selection strategies, columns are aggregation strategies. We highlighted the \fbox{best} and \underline{second best} one in each column. The average shows that our client selection strategy works better than other selection strategies when combined with a variety of aggregation methods.}
  \label{tab:combination}
\end{table*}

We analyze \confed's performance under different conditions, focusing on datasets, heterogeneity modeling, and comparative analysis with baselines. In all experiments, we report the \emph{early-stop accuracy} achieved during federated optimization.

\begin{figure*}[ht]
\centering
~\hspace{-15pt}
\subfloat[]{\begin{tikzpicture}
\node at (-3.2,-.4)[anchor=west]{\includegraphics[width=0.25\textwidth]{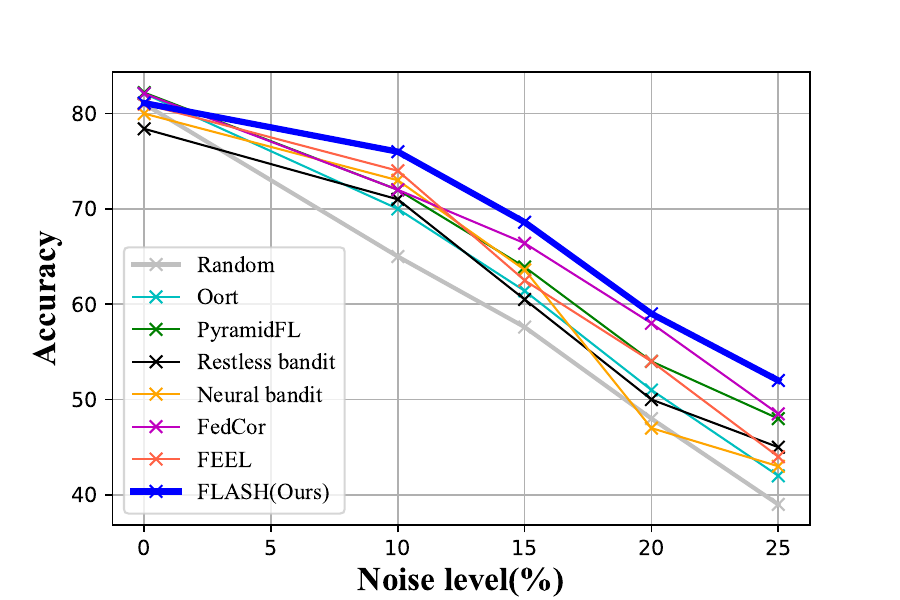}};
\end{tikzpicture}
}
~\hspace{-15pt}
\subfloat[]{\begin{tikzpicture}
\node at (-3.2,-.4)[anchor=west] {\includegraphics[width=0.25\textwidth]{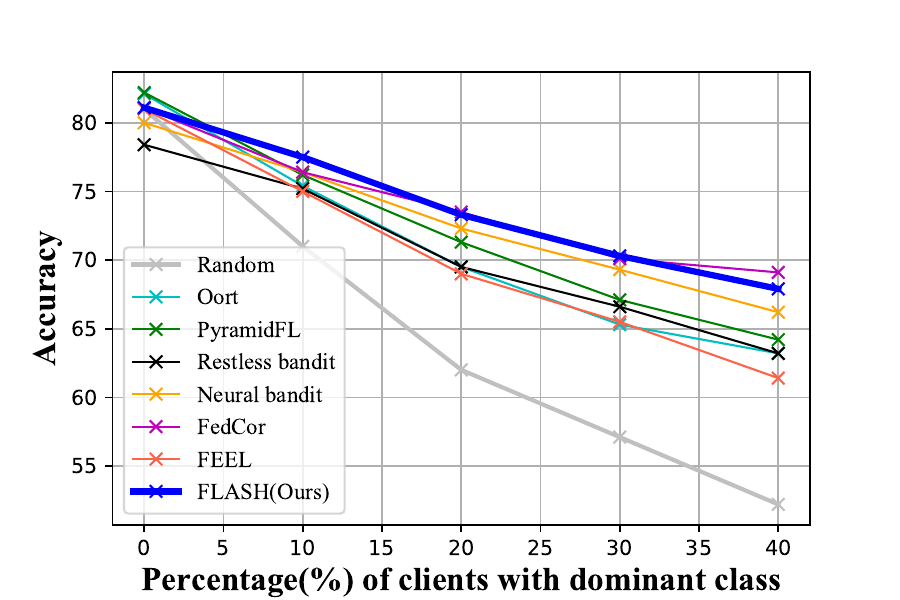}};
\end{tikzpicture}
}
~\hspace{-15pt}
\subfloat[]{\begin{tikzpicture}
\node at (-3.2,-.4)[anchor=west]{\includegraphics[width=0.25\textwidth]{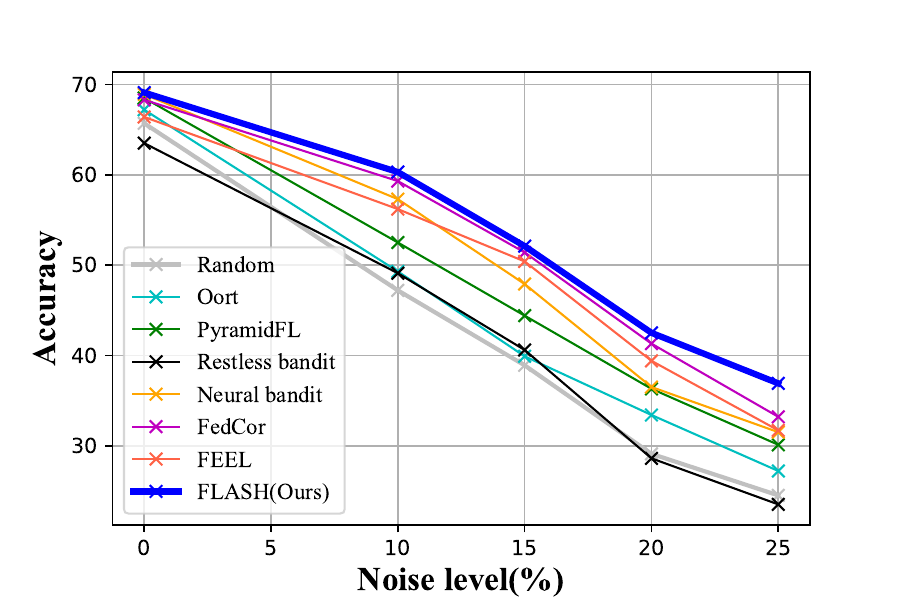}};
\end{tikzpicture}
}
~\hspace{-15pt}
\subfloat[]{\begin{tikzpicture}
\node at (-3.2,-.4)[anchor=west] {\includegraphics[width=0.25\textwidth]{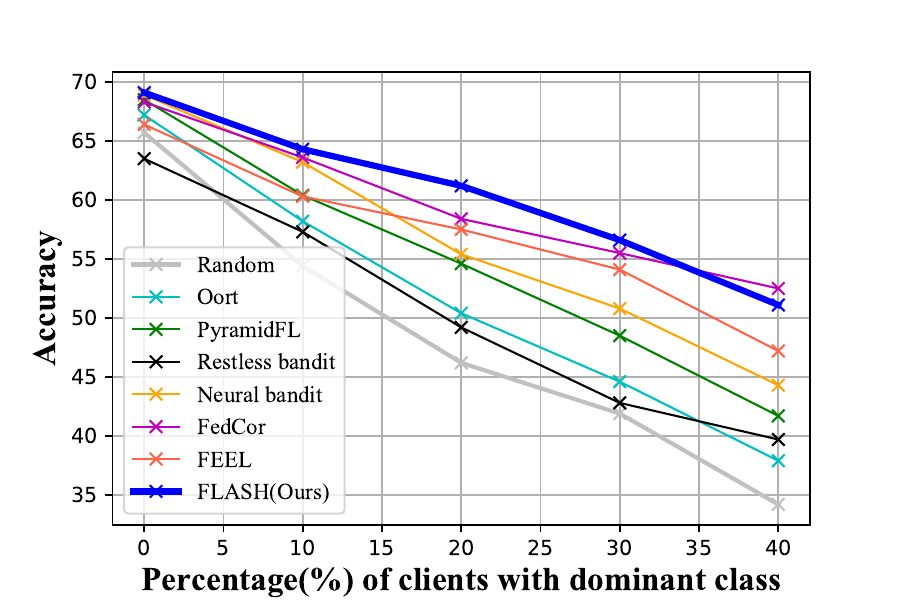}};
\end{tikzpicture}
}
\vspace*{-0.2cm}
\caption[caption]{Best global model test accuracy on CIFAR10 (a-b) and FEMNIST (c-d) dataset for different selection algorithms (with FedAvg aggregation) as the noise level (a,c), and non-IIDness of the data distribution (a,d) are varied.}

\label{fig:singlehetero}
\end{figure*}

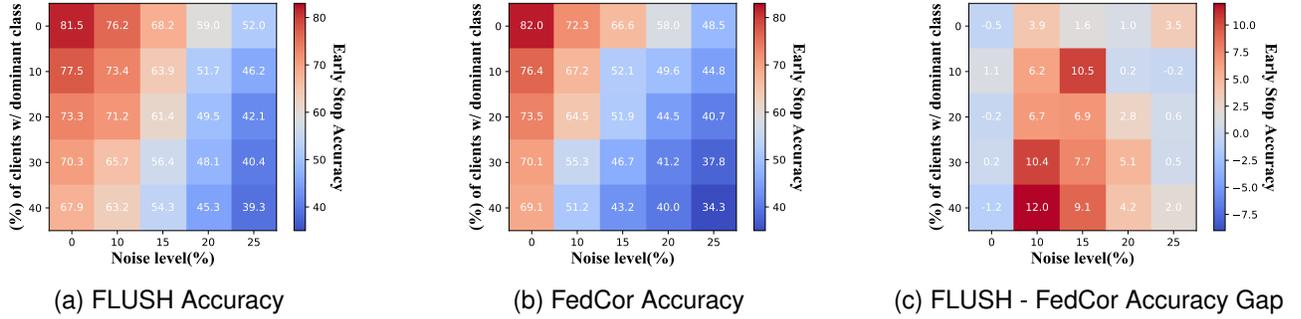
\begin{figure*}[!ht]

\centering
~\hspace{-10pt}
\subfloat[\confed Accuracy]{\begin{tikzpicture}
\node at (-3.2,-.4)[anchor=west]{\includegraphics[width=0.33\textwidth]{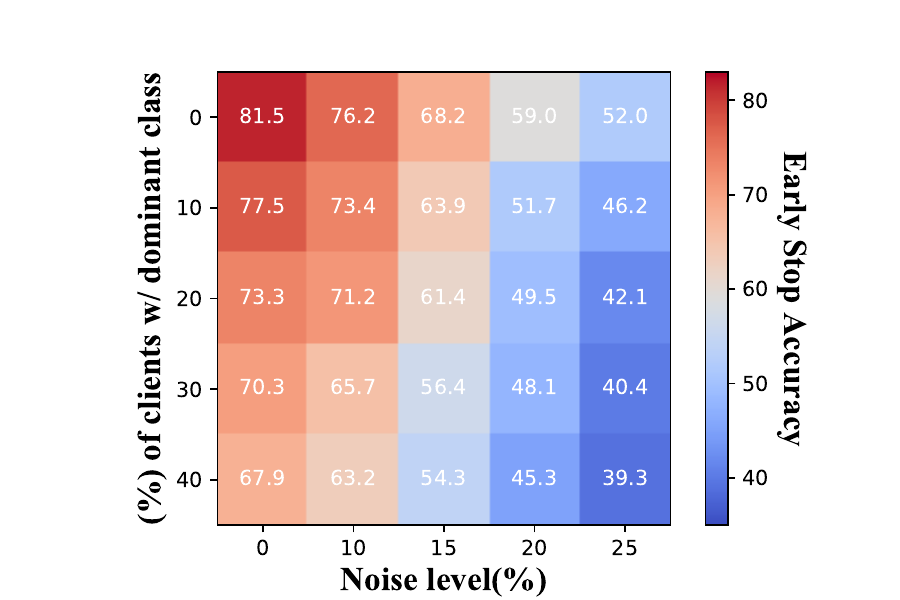}};
\node at (-1.1,1.35)[anchor=west][scale=0.6,opacity=1]{};
\node at (-2.9,-1.6)[anchor=west][scale=0.6,rotate=90,opacity=1]{ };
\end{tikzpicture}
}
~\hspace{-10pt}
\subfloat[FedCor Accuracy]{\begin{tikzpicture}
\node at (-3.2,-.4)[anchor=west] {\includegraphics[width=0.33\textwidth]{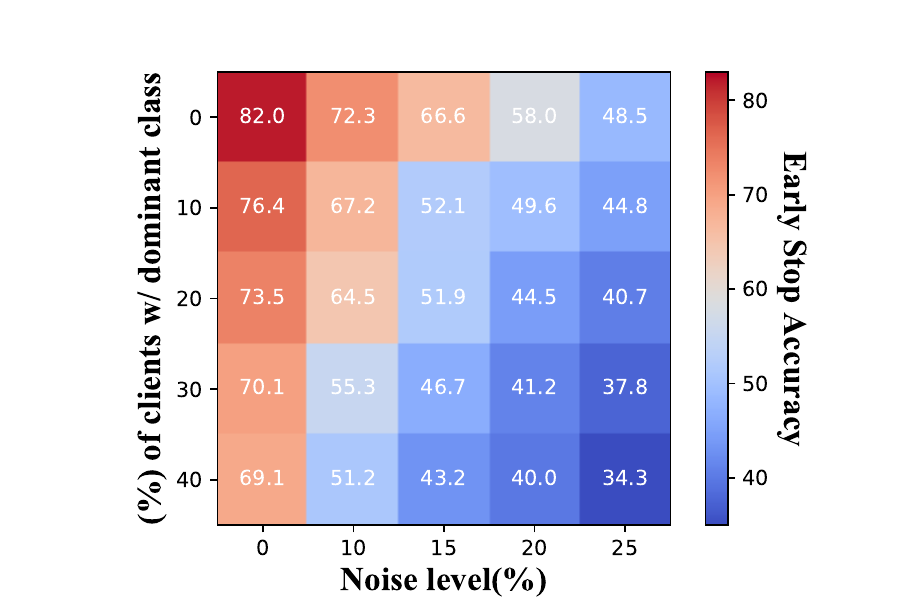}};
\node at (-0.7,1.35)[anchor=west][scale=0.6,opacity=1]{};
\end{tikzpicture}
}
~\hspace{-10pt}
\subfloat[\confed~  - FedCor Accuracy Gap]{\begin{tikzpicture}
\node at (-3.2,-.4)[anchor=west]{\includegraphics[width=0.33\textwidth]{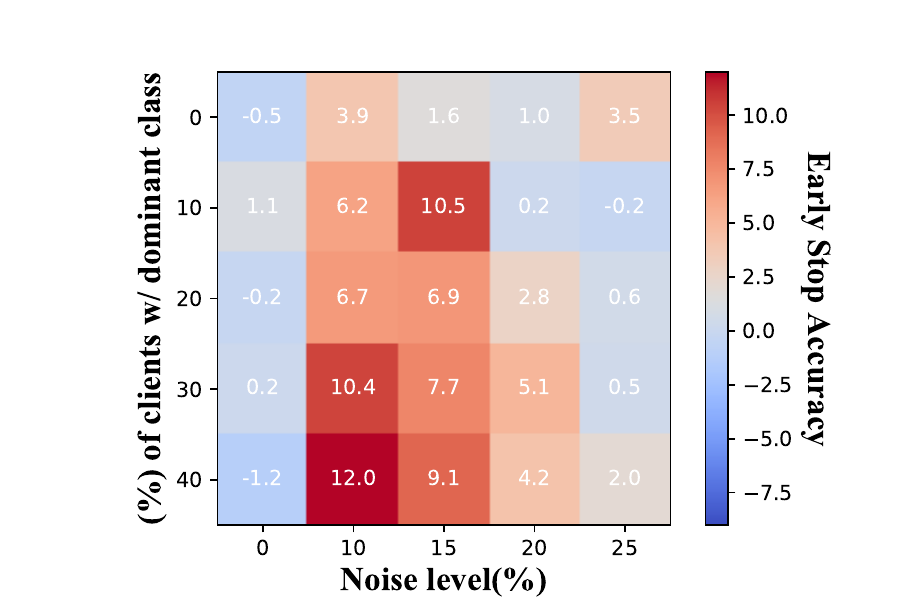}};
\node at (-0.7,1.35)[anchor=west][scale=0.6,opacity=1]{};
\end{tikzpicture}
}
\caption[caption]{
Heatmaps demostrate the best test accuracy that \confed~and FedCor (state-of-the-art) can achieve under varying levels of combination of two heterogeneities with FedAvg aggregation: (a) \confed, (b) FedCor, and (c) \confed-FedCor. The larger
the area of the red and orange regions, the better the corresponding algorithm performs on more heterogeneities. The advantage of \confed over FedCor is more visible when the problem involves both label noise and non-IIDness. The improvement is 3.76\% improvement on average over all noise/non-IID levels and can be more than 10\%.
}
\vspace{-0.17in}
\label{fig:multihetero}
\end{figure*}

\subsection{Datasets, Heterogeneity Models, and Baselines}
\label{sec:datasets}
\textbf{Datasets.} We evaluate \confed~on two widely-used FL datasets: CIFAR-10 \cite{krizhevsky2009learning} with 60,000 32x32 color images in 10 classes, and FEMNIST \cite{caldas2018leaf}, a hand-written digits dataset with 62 classes built by partitioning Extended MNIST based on writers.\\
\textbf{Modeling Heterogeneity.} We model three types of heterogeneity:
1) Non-IIDness: For heterogeneous clients, 80\% of data comes from a single class while 20\% from other classes; homogeneous clients maintain uniform distribution.
2) Label noise: Noise levels follow a Beta distribution $B(\alpha_{Beta}, \beta_{Beta})$ with $\alpha_{Beta}$ in $\{5,10,15,20,25\}$ and $\beta_{Beta}=100-\alpha_{Beta}$, following \cite{ortego2021towards,albert2022addressing}.
3) Latency: Device execution times follow a shifted exponential distribution \cite{shi2020device}.\\
For non-IID modeling, we assign a pronounced data skew (e.g., 80\% from one class) to a controlled percentage of clients (e.g., "30\% non-IIDness" explicitly means 30\% of clients exhibit this 80/20 profile). This method provides direct, intuitive control over the proportion of clients with significant skew\cite{ortego2021towards}, which is advantageous for studying the direct impact on client selection algorithm performance as this quantity of heterogeneity varies. While the Dirichlet distribution ($\text{dir}(\alpha)$) is a common alternative, controlling the exact proportion of clients with a specific high degree of skew (e.g., a dominant class holding >80\% of data) is less direct and requires precise $\alpha$ tuning. For instance, numerical simulations indicate that achieving approximately 40\% dominant-class clients requires $\alpha \approx 0.06-0.07$, while $\alpha=0.01$ yields about 89\% such clients, and $\alpha=0.10$ results in roughly 12\%. Our experiments, which examine scenarios with up to 40\% skewed clients, therefore correspond to challenging non-IID conditions (equivalent to $\text{dir}(\alpha)$ with $\alpha < 0.1$), confirming a rigorous evaluation environment.\\
\textbf{Baseline methods.} We compare \confed~with combinations of recent client selection strategies and aggregation methods. Selection methods focus on client contribution \cite{ren2020scheduling,cao2022birds}, client drift \cite{FedCor}, and communication delay \cite{lai2021oort,li2022pyramidfl}. Aggregation methods aim to optimize model aggregation \cite{FedDF, wang2020tackling, karimireddy2020scaffold} and reduce client drift \cite{FedProx, FedBiO, RHFL}.

\begin{figure*}[!ht]

\centering
~\hspace{-15pt}
\subfloat[]{\begin{tikzpicture}
\node at (-3.2,-.4)[anchor=west]{\includegraphics[width=0.25\textwidth]{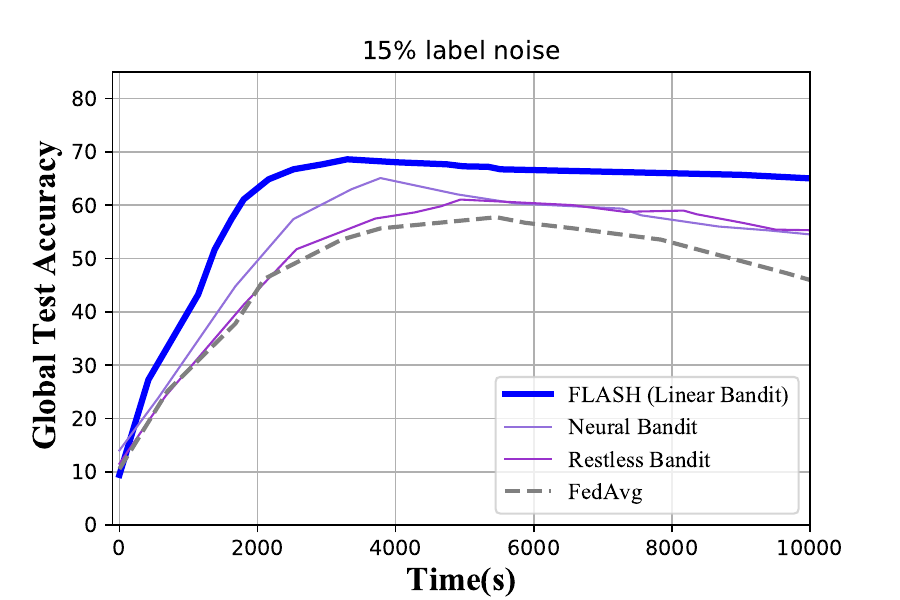}};
\end{tikzpicture}
}
~\hspace{-15pt}
\subfloat[]{\begin{tikzpicture}
\node at (-3.2,-.4)[anchor=west] {\includegraphics[width=0.25\textwidth]{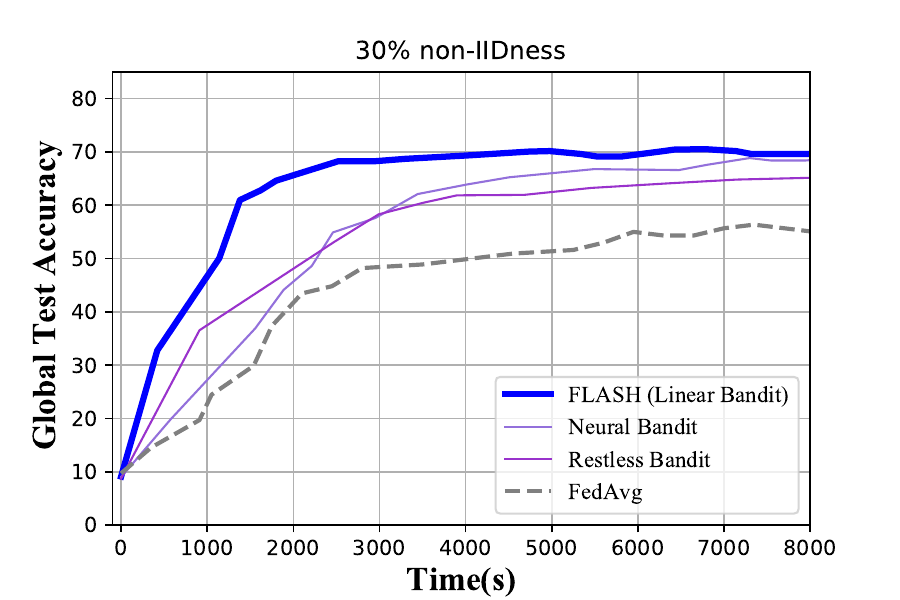}};
\end{tikzpicture}
}
~\hspace{-15pt}
\subfloat[]{\begin{tikzpicture}
\node at (-3.2,-.4)[anchor=west]{\includegraphics[width=0.25\textwidth]{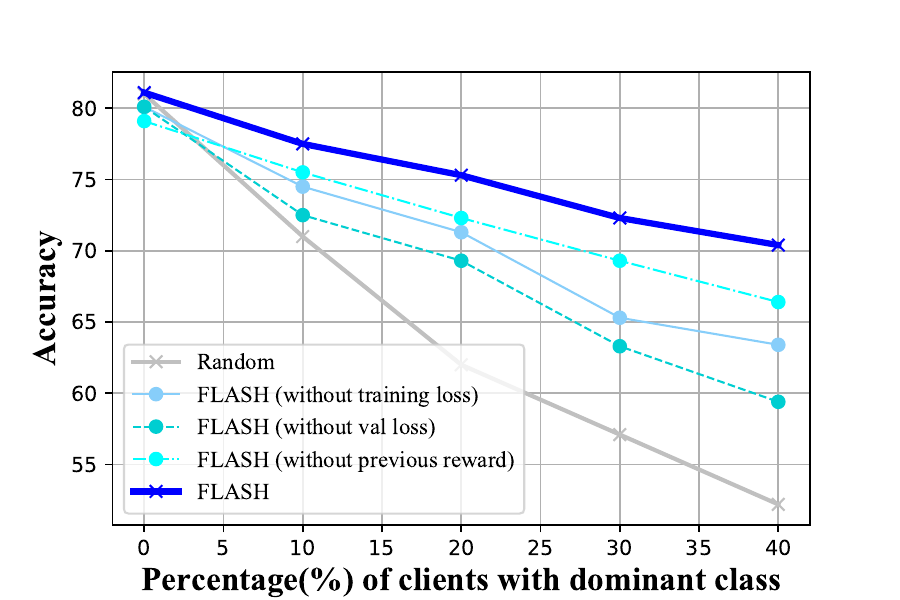}};
\end{tikzpicture}
}
~\hspace{-15pt}
\subfloat[]{\begin{tikzpicture}
\node at (-3.2,-.4)[anchor=west]{\includegraphics[width=0.25\textwidth]{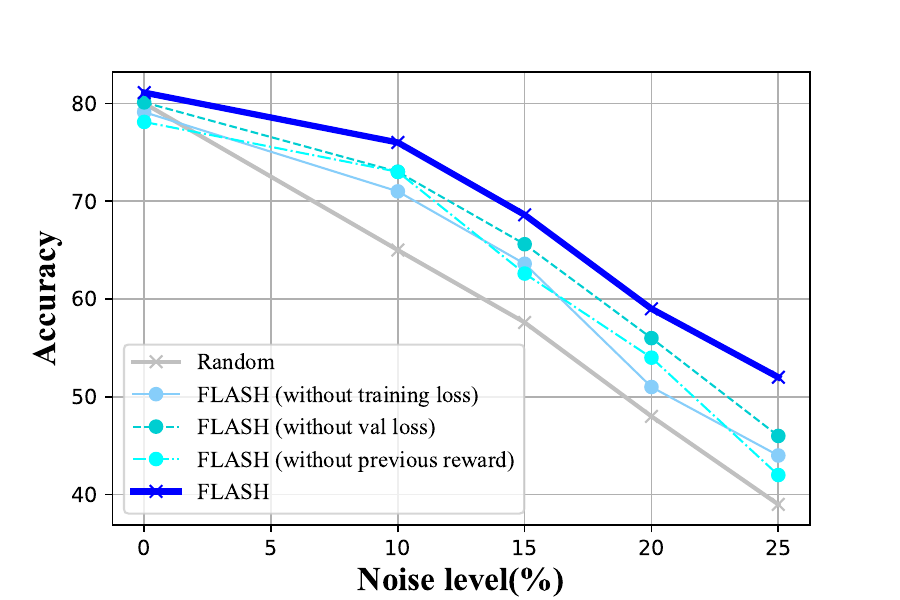}};
\end{tikzpicture}
}

\vspace{-2pt}\caption[caption]{(a-b) Test accuracy for \confed~(with FedAvg) applying different types of multi-armed bandits under different settings of heterogeneity on CIFAR dataset: (a) under noisy setting (b) under non-IID setting. These figures depict the training time required and the achieved global model accuracy when replacing \confed's linear bandit with other types of bandit. It is clear that \confed~ achieves the same accuracy with far less training time. (c-d) Ablation studies of the context vector elements of \confed~(with FedAvg) on CIFAR: Best global model test accuracy as the (c) non-IIDness, (d) noise level of the data distribution is varied. These figures illustrate the potential performance degradation of the global model when specific context vector elements are removed. } 
\vspace{-0.17in}
\label{fig:ablation_and_bandit}
\end{figure*}

\begin{figure*}[!ht]

\centering
~\hspace{-15pt}
\subfloat[CIFAR10]{\begin{tikzpicture}
\node at (-3.2,-.4)[anchor=west]{\includegraphics[width=0.20\textwidth]{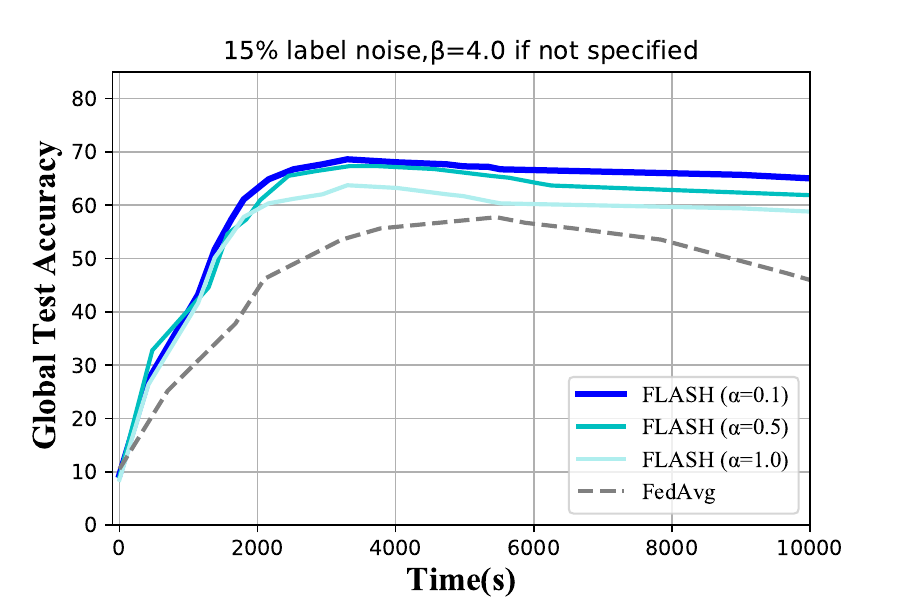}};
\end{tikzpicture}
}
~\hspace{-15pt}
\subfloat[CIFAR10]{\begin{tikzpicture}
\node at (-3.2,-.4)[anchor=west] {\includegraphics[width=0.20\textwidth]{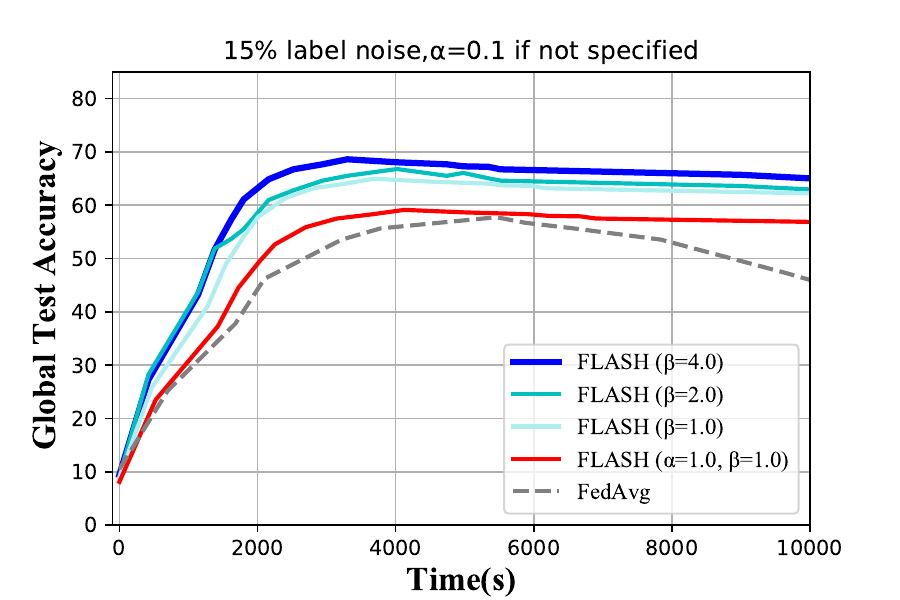}};
\end{tikzpicture}
}
~\hspace{-15pt}
\subfloat[FEMNIST]{\begin{tikzpicture}
\node at (-3.2,-.4)[anchor=west]{\includegraphics[width=0.20\textwidth]{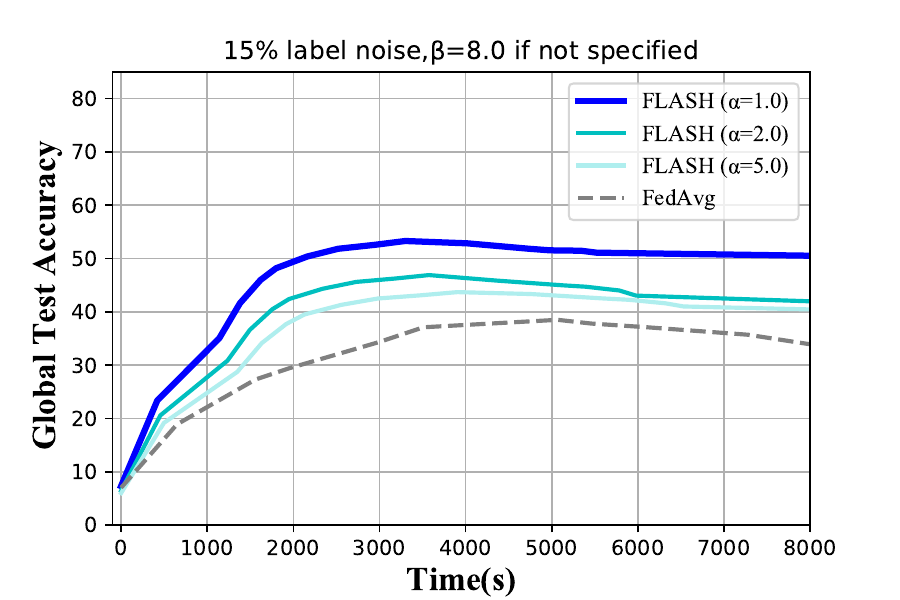}};
\end{tikzpicture}
}
~\hspace{-15pt}
\subfloat[FEMNIST]{\begin{tikzpicture}
\node at (-3.2,-.4)[anchor=west]{\includegraphics[width=0.20\textwidth]{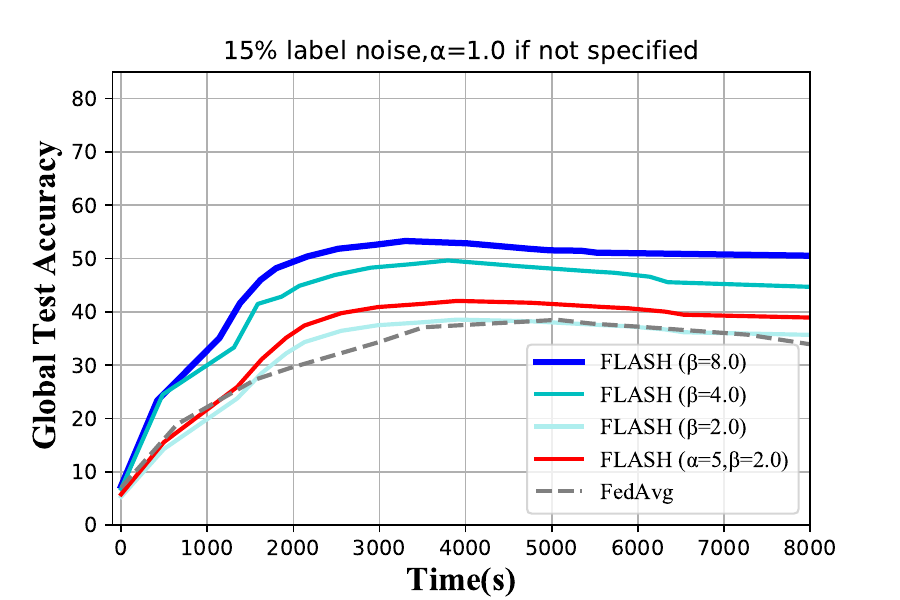}};
\end{tikzpicture}
}
\vspace{-2pt}\caption[caption]{Test accuracy for \confed (with FedAvg), applying different combinations of $(\alpha, \beta)$ in $\mathcal{L}_{robust}$.$A=-4$ is fixed for all the dataset: (a-b) CIFAR10, (c-d) FEMNIST. The choice of $(\alpha,\beta)$ are: (a) $\alpha=\{0.1,0.5,1.0\},\beta=4.0$, (b) $\alpha=0.1,\beta=\{4.0,2.0,1.0\}$ and $(\alpha,\beta)=(1.0,1.0)$,(c) $\alpha=\{0.1,0.5,1.0\},\beta=8.0$, (d) $\alpha=1.0,\beta=\{8.0,4.0,2.0\}$ and $(\alpha,\beta)=(5,1.0)$}
\vspace{-0.17in}
\label{fig:alpha_beta}
\end{figure*}
\subsection{Implementation Details}
We use $m=50$ clients for CIFAR-10 and $m=3550$ for FEMNIST, selecting $M_t=0.2\cdot m$ clients per round for maximum $n=1500$ rounds. Algorithm parameters are set to $\delta=5\times 10^{-2}$ and $\lambda=1$. All clients use ResNet-18 trained locally for 5 epochs with Adam optimizer.\\
For $\mathcal{L}_{robust}$, we fix $A=-4$ and tune $(\alpha, \beta)$. A moderately large $\alpha$ is recommended for challenging datasets, while $\beta$'s impact varies with dataset complexity (see Fig.~\ref{fig:alpha_beta}). For latency simulation, we model duration as $\tau_t = \max_{i\in S_t}\{T_i\}$ where $T_i - \alpha_T N_i \sim \text{Exp}(1/\lambda_T N_i)$, with $\alpha_T=1$ and $\lambda_T$ varying in $\{1, 10, 100\}$.

\subsection{Results and Analysis}
\label{sec:accuracy}
\textbf{Generalizability:} Table~\ref{tab:combination} shows \confed~provides better average performance across different aggregation strategies. While some methods like FedCor~\cite{FedCor} excel with specific strategies, \confed~demonstrates consistent performance across various conditions.\\
\textbf{Heterogeneity Analysis:} Figures \ref{fig:singlehetero} and \ref{fig:multihetero} demonstrate \confed's effectiveness under various heterogeneity levels. While some algorithms may outperform \confed~in specific scenarios, \confed~shows superior performance with multiple concurrent heterogeneities.\\
\textbf{Bandit Algorithm Comparison:} Fig.~\ref{fig:ablation_and_bandit} (a,b) show that more complex MAB algorithms like Neural Bandit \cite{cao2022birds} or Restless Bandit \cite{yemini2021restless} don't improve performance but increase computational overhead, indicating the feasibility and optimality of our linear bandit application.\\
\textbf{Context Vector Analysis:} Our ablation studies (Fig.~\ref{fig:ablation_and_bandit}, (c,d)) reveal that local training loss is crucial for noisy datasets, while local validation loss is more important in non-IID settings. Previous round reward contributes to stable improvement across various settings. The "Duration" feature effectively addresses latency heterogeneity.\\
\textbf{Different combinations of $(\alpha, \beta, A)$ in $\mathcal{L}_{robust}$:} The RCE term in the loss (Eqn. 5) can be further simplified:
$$\begin{aligned}
\mathcal{L}_{RCE}(\vct{a}, y)&=-\sum_{k=1}^K [p(\vct{a}; \vct{w}^t_i)]_k \log [y]_k\\
=-&A\sum_{k\neq y}[p(\vct{a}; \vct{w}^t_i)]_k=-A(1-[p(\vct{a}; \vct{w}^t_i)]_{k=y}).
\end{aligned}$$
Since tuning $A$ is equivalent to scaling $\beta$, we fix $A=-4$ and only tune hyperparameters $(\alpha,\beta)$. For parameter tuning in $\mathcal{L}_{robust}$, both $\alpha$ and $\beta$ require careful consideration:\\
\textit{Parameter $\alpha$:}
Large $\alpha$ leads to overfitting, while small $\alpha$ reduces overfitting in $\mathcal{L}_{CE}$ but slows convergence (similar to using only $\mathcal{L}_{RCE}$). A moderately large $\alpha$ is recommended, especially for challenging datasets like FEMNIST.\\
\textit{Parameter $\beta$:}
Its impact depends on dataset complexity and $\alpha$ selection: For simple datasets (e.g., CIFAR10) with well-chosen $\alpha$: $\beta$ has minimal impact on training (shown for $\beta=1,2,4$ in Fig.~\ref{fig:alpha_beta}(b)). For challenging datasets (e.g., FEMNIST) or poorly-chosen $\alpha$: results are highly sensitive to $\beta$ choice, as demonstrated in Fig.~\ref{fig:alpha_beta}(b) and (d). 
\section{Conclusions}
We have addressed an open, but critically important, problem in federated learning, namely how to simultaneously deal with multiple kinds of heterogeneities that arise across local clients. These include latencies, noisy labels at the clients, and varying data distributions across the clients. We proposed \confed~ --  a flexible client selection algorithm that automatically incorporates rich contextual information associated with the heterogeneity at the clients via contextual multi-armed bandits. 
On two of the most commonly-used datasets, \confed~shows significant performance improvements over existing client selection methods, especially when multiple heterogeneities are present simultaneously. Moreover, we showed the generalizability of \confed~when combined with a variety of global aggregation methods.


\bibliographystyle{IEEEbib}
\bibliography{refs.bib}
\end{document}